\documentclass[runningheads]{llncs}
\usepackage{color}
\usepackage{tikz}

\usetikzlibrary{shapes,decorations,arrows,calc,arrows.meta,fit,positioning}
\tikzset{
    -Latex,auto,node distance =1 cm and 1 cm,semithick,
    state/.style ={ellipse, draw, minimum width = 0.7 cm},
    point/.style = {circle, draw, inner sep=0.04cm,fill,node contents={}},
    bidirected/.style={Latex-Latex,dashed},
    el/.style = {inner sep=2pt, align=left, sloped}
}

\usepackage[T1]{fontenc}
\usepackage{graphicx}
\usepackage{capt-of}  
\usepackage{cuted}
\usepackage[backend=biber, sorting=none, giveninits=true]{biblatex}
\usepackage{graphicx}%
\usepackage{multirow}%
\usepackage{amsmath,amssymb,amsfonts}%
\usepackage{mathrsfs}%
\usepackage[figuresright]{rotating}%
\usepackage{xcolor}%
\usepackage{textcomp}%
\usepackage{manyfoot}%
\usepackage{booktabs}%
\usepackage{algorithm}%
\usepackage{algorithmicx}%
\usepackage{algpseudocode}%
\usepackage{program}%
\usepackage{listings}%
\usepackage{wrapfig}
\usepackage{vruler}
\usepackage{setspace}
\usepackage{comment}
\usepackage{siunitx}
\usepackage{booktabs}

\usepackage{multicol}

\begin{document}
%
\title{Emergent Causality \\ and the Foundation of Consciousness}

%
%
\author{Michael Timothy Bennett\inst{1} \\ \orcidID{0000-0001-6895-8782} 
}
\authorrunning{Michael Timothy Bennett}
%
\institute{The Australian National University  \\\email{michael.bennett@anu.edu.au} \\\url{http://www.michaeltimothybennett.com/}
}
\maketitle              

\begin{abstract}
To make accurate inferences in an interactive setting, an agent must not confuse passive observation of events with having intervened to cause them. The $do$ operator formalises interventions so that we may reason about their effect. Yet there exist pareto optimal mathematical formalisms of general intelligence in an interactive setting which, presupposing no explicit representation of intervention, make maximally accurate inferences. We examine one such formalism. We show that in the absence of a $do$ operator, an intervention can be represented by a variable. We then argue that variables are abstractions, and that need to explicitly represent interventions in advance arises only because we presuppose these sorts of abstractions. The aforementioned formalism avoids this and so, initial conditions permitting, representations of relevant causal interventions will emerge through induction. 
These emergent abstractions function as representations of one’s self and of any other object, inasmuch as the interventions of those objects impact the satisfaction of goals. We argue that this explains how one might reason about one's own identity and intent, those of others, of one's own as perceived by others and so on. In a narrow sense this describes what it is to be aware, and is a mechanistic explanation of aspects of consciousness\footnote{Appendices are to be found on GitHub \cite{bennett2023appendices}.}.
\end{abstract}

\section{Introduction}
\label{intro}
An agent that interacts in the world cannot make accurate inferences unless it distinguishes the passive observation of an event from it having intervened to cause that event \cite{pearl2018,ortega2021}. Say we had two variables $R,C \in \{true,false\}$, where:
$$C = true \leftrightarrow \textit{``Larry put on a raincoat''} \text{ and } R = true \leftrightarrow  \textit{``It rained''}$$
Assume we have seen it rain only when Larry had his raincoat on, and he has only been seen in his raincoat during periods of rain. Based on these observations, the conditional probability of it raining if Larry is wearing his raincoat is $p(R=true \mid C=true) = 1$. A naive interpretation of this is that we can make it rain\newpage \noindent by forcing Larry to wear a raincoat, which is absurd. When we intervene to make Larry wear a raincoat, the event that takes place is not \textit{``Larry put on a raincoat''} but actually \textit{``Larry put on a raincoat because we forced him to''}. It is not that Bayesian probability is wrong, but interactivity complicates matters. By intervening we are acting upon the system from the outside, to disconnect those factors influencing the choice of clothing. 
The ``do'' operator \cite{pearl1995,pearl2009} resolves this in that $do[C = true]$ represents the intervention. It allows us to express notions such as $p(R=true \mid do[C=true]) = p(R=true) \neq p(R=true \mid C=true) = 1$, which is to say that intervening to force Larry to wear a raincoat has no effect on the probability of rain, but passively observing Larry put on a raincoat still indicates rain with probability $1$.
To paraphrase Judea Pearl, one variable causes another if the latter listens for the former \cite{pearl2018}. The variable $R$ does not listen to the $C$. $C$ however does listen to $R$, meaning to identify cause and effect imposes a hierarchy on one's representation of the world (usually represented with a directed acyclic graph).
This suggests that, if accurate inductive inference is desired, we must presuppose something akin to the $do$ operator. 
Yet there exist pareto optimal mathematical formalisms of general intelligence in an interactive setting which, given no explicit representation of intervention, make maximally accurate inferences \cite{hutter2010,bennett2022a,bennett2023appendices}. 
Given that the distinction between observation and intervention is necessary to make accurate inductive inferences in an interactive setting, this might seem to present us with a contradiction. One cannot accurately infer an equivalent of the $do$ operator if such a thing is a necessary precondition of accurate inductive inference. 
We resolve this first by showing that we can substitute an explicit $do$ operator with variables representing each intervention. Then, using one of the aforementioned formalisms, we argue that need to explicitly represent intervention as a variable only arises if we presuppose abstractions \cite{harnad1990} like variables. If induction does not depend upon abstractions as given, then abstractions representing interventions may emerge through inductive inference. 
Beyond distinguishing passive observation from the consequences of one's own interventions, these emergent abstractions can also distinguish between the interventions and observations of others. This necessitates the construction of abstract identities and intents. We suggest this is a mechanistic explanation of awareness, in a narrow sense of the term. By narrow we mean functional, access, and phenomenal consciousness, and only if the latter is defined as ``first person functional consciousness'' \cite{franklin2008,boltuc2012}; recognising phenomenal content such as light, sound and movement with one's body at the centre of it all \cite{block2002}. To limit scope, we do not address ``the hard problem'' \cite{chalmers1995}.

\section{Additional background}
This section introduces relevant background material. The reader may wish to skip ahead to section $3$ and refer here as needed. In recognition of the philosophical nature of this topic we present arguments rather than mathematical proofs, and the paper should be understandable without delving too deeply into the math.
While all relevant definitions are given here, context is provided by the papers in which these definitions originated, and in technical appendices available on GitHub \cite{bennett2023appendices}.
To those more familiar with the agent environment paradigm, how exactly these definitions formalise cognition may seem unclear. Neither agent nor environment are defined. This is because it is a formalism of enactivism \cite{ward2017}, which holds that cognition extends into and is enacted within the environment. What then constitutes the agent is unclear. In light of this, and in the absence of any need to define an agent absent an environment, why preserve the distinction? Subsequently, the agent and environment are merged to form a task \cite{bennett2022a}, which may be understood as context specific manifestations of intent, or snapshots of what bears some resemblance to ``Being-in-the-world'' as described by Heidegger \cite{heidegger2020}. In simpler terms, this reduces cognition to a finite set of decision problems \cite{bennett2022a}. One infers a model from past interactions, and then makes a decision based upon that model (akin to a supervised learner fitting a function to labelled data, then using that to generate labels for unlabelled data). Arguments as to why only finite sets are relevant are given elsewhere \cite[p. 2]{bennettmaruyama2021b}. 

\subsection{List of definitions}
Refer to the appendices \cite{bennett2023appendices} and the related papers \cite{bennett2023a,bennett2023b,bennett2023d} for further information regarding these definitions.

\begin{definition}[environment]\label{d1}
\begin{itemize}{
    \item  We assume a set $\Phi$ whose elements we call \textbf{states}, one of which we single out as the \textbf{present state}.
    \item A \textbf{declarative program} is a function $f : \Phi \rightarrow \{true, false\}$, and we write $P$ for the set of all declarative programs. By an \textbf{objective truth} about a state $\phi$, we mean a declarative program $f$ such that $f(\phi) = true$.
    }
\end{itemize}
\end{definition}
\begin{definition}[implementable language] \label{d2}
\begin{itemize}{
    \item $\mathfrak{V} = \{V \subset P : V \ is \ finite\}$ is a set whose elements we call \textbf{vocabularies}, one of which\footnote{The vocabulary $\mathfrak{v}$ we single out represents the sensorimotor circuitry with which an organism enacts cognition - their brain, body, local environment and so forth.} we single out as \textbf{the vocabulary} $\mathfrak{v}$ for an implementable language.
    \item ${L_\mathfrak{v}} = \{ l \subseteq \mathfrak{v} : \exists \phi \in \Phi \ (\forall p \in l : p(\phi) = true) \}$ is a set whose elements we call \textbf{statements}. $L_\mathfrak{v}$ follows from $\Phi$ and $\mathfrak{v}$. We call $L_\mathfrak{v}$ an \textbf{implementable language}.
    \item  $l \in {L_\mathfrak{v}}$ is \textbf{true} iff the present state is $\phi$ and $\forall p \in l : p(\phi) = true$.
    \item The \textbf{extension of a statement} $a \in {L_\mathfrak{v}}$ is $Z_a = \{b \in {L_\mathfrak{v}} : a \subseteq b\}$.
    \item The \textbf{extension of a set of statements} $A \subseteq {L_\mathfrak{v}}$ is $Z_A = \bigcup\limits_{a \in A} Z_a$.
    }
\end{itemize}

\noindent {\normalfont(Notation)} $Z$ with a subscript is the extension of the subscript\footnote{e.g. $Z_s$ is the extension of $s$.}.

\end{definition}

\begin{definition}[{$\mathfrak{v}$}-task]\label{d3} For a chosen $\mathfrak{v}$, a task $\alpha$ is $\langle {S}_\alpha, {D}_\alpha, {M}_\alpha \rangle$ where:\begin{itemize}{ 
    \item ${S}_\alpha \subset L_\mathfrak{v}$ is a set whose elements we call \textbf{situations} of $\alpha$.
    \item ${S_\alpha}$ has the extension $Z_{S_\alpha}$, whose elements we call \textbf{decisions} of $\alpha$. 
    \item ${D_\alpha} = \{z \in Z_{S_\alpha} : z \ is \ correct \}$ is the set of all decisions which complete $\alpha$. 
    \item ${M_\alpha} = \{l \in L_\mathfrak{v} : {Z}_{S_\alpha} \cap Z_{l} = {D_\alpha}\}$ whose elements we call \textbf{models} of $\alpha$.}
\end{itemize}
$\Gamma_\mathfrak{v}$ is the set of all tasks for our chosen $\mathfrak{v} \in \mathfrak{V}$.\\

\noindent{\normalfont(Notation)} If $\omega \in \Gamma_\mathfrak{v}$, then we will use subscript $\omega$ to signify parts of $\omega$, meaning one should assume $\omega = \langle {S}_\omega, {D}_\omega, {M}_\omega \rangle$ even if that isn't written.\\

\noindent {\normalfont(How a task is completed)} Assume we've a $\mathfrak{v}$-task $\omega$ and a hypothesis $\textbf{h} \in L_\mathfrak{v}$ s.t.\begin{enumerate}{
    \item we are presented with a situation ${s} \in {S}_\omega$, and
    \item we must select a decision $z \in Z_{s} \cap Z_\textbf{h}$.
    \item If $z \in {D}_\omega$, then $z$ is correct and the task is complete. This occurs if $\textbf{h} \in {M}_\omega$.}
\end{enumerate} 
\end{definition}

\begin{definition}[probability]\label{d4} We assume a uniform distribution over $\Gamma_\mathfrak{v}$.
\end{definition}

\begin{definition}[generalisation]\label{d5} A statement $l$ generalises to $\alpha \in \Gamma_\mathfrak{v}$ iff $l \in M_\alpha$. We say $l$ generalises from $\alpha$ to  $\mathfrak{v}$-task $\omega$ if we first obtain ${l}$ from ${M}_\alpha$ and then find it generalises to $\omega$.
\end{definition}

\begin{definition}[child and parent]\label{d6} A $\mathfrak{v}$-task $\alpha$ is a child of $\mathfrak{v}$-task $\omega$ if ${S}_\alpha \subset {S}_\omega$ and ${D}_\alpha \subseteq {D}_\omega$. 
This is written as $\alpha \sqsubset \omega$. If $\alpha \sqsubset \omega$ then $\omega$ is then a parent of $\alpha$.
\end{definition}

\begin{definition}[weakness]\label{d7} The weakness of $l \in L_\mathfrak{v}$ is $\lvert Z_{l} \rvert$.
\end{definition}

\begin{definition}[induction]\label{d8} $\alpha$ and $\omega$ are $\mathfrak{v}$-tasks such that $\alpha \sqsubset \omega$. Assume we are given a proxy $q_\mathfrak{v} \in Q$, the complete definition of $\alpha$ and the knowledge that $\alpha \sqsubset \omega$. We are not given the definition of $\omega$. The process of induction would proceed as follows:
\begin{enumerate}{
    \item Obtain a hypothesis by computing a model $\mathbf{h} \in \underset{{m} \in {M}_\alpha}{\arg\max} \ q_\mathfrak{v}(m)$.
    \item If $\mathbf{h} \in {M}_\omega$, then we have generalised from $\alpha$ to $\omega$.}
\end{enumerate}
\end{definition}

\subsection{Premises}
For the purpose of argument we will adopt the following premises:
\begin{quote}\label{p2} \textbf{(prem. 1)} To maximise the probability that induction generalises from $\alpha$ to $\omega$, it is necessary and sufficient to maximise weakness. \cite{bennett2023appendices}\end{quote}
For our argument this optimality is less important than the representation of interventions it implies. In any case the utility of weakness as a proxy is not limited to lossless representations or optimal performance. Approximation may be achieved by selectively forgetting outliers\footnote{For example, were we trying to generalise from $\alpha$ to $\omega$ (where $\alpha \sqsubset \omega$) and knew the definition of $\alpha$ contained misleading errors, we might selectively forget outlying decisions in $\alpha$ to create a child $\gamma = \langle S_\gamma, D_\gamma, M_\gamma \rangle$ (where $\gamma \sqsubset \alpha$) such that $M_\gamma$ contained far weaker hypotheses than $M_\alpha$.}, a parallel to how selective amnesia \cite{bekinschtein2018} can help humans reduce the world to simple dichotomies \cite{berlin1990} or confirm preconceptions \cite{nickerson1998}. Likewise, a task expresses a threshold beyond which decisions are ``good enough'' \cite{bennettmaruyama2022a}. The proof of optimality merely establishes the upper bound for generalisation. 
As a second premise, we shall require the emergence or presupposition of representations of interventions:
\begin{quote}\label{p3} \textbf{(prem. 2)} To make accurate inductive inferences in an interactive setting, an agent must not confuse the passive observation of an event with having intervened to cause that event. \cite{pearl2018}\end{quote}

\section{Emergent Causality}
The formalism does not presuppose an operator representing intervention. Given our premises, we must conclude from this that either that \textbf{(prem. 1)} is false, or induction as in definition \ref{d8} will distinguish passive observation of an event from having intervened to cause that event.

\subsection{The $do$ operator as a variable in disguise}
In the introduction we discussed an example involving binary variables $R$ (rain) and $C$ (raincoat). From $p(R=true \mid C= true) = 1$ we drew the absurd conclusion that if we intervene to make $C=true$, we can make it rain. The true relationship between $R$ and $C$ is explained by a directed acyclic graph:
$$
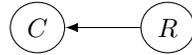
\begin{tikzpicture}
    \node[state] (c) at (0,0) {$C$};
    \node[state] (r) [right =of c] {$R$};

    \path (r) edge (c);

\end{tikzpicture}
$$
The intervention $do[C=c]$ deletes an edge (because rain can have no effect on the presence of a coat we've already forced Larry to wear) giving the following:
$$
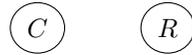
\begin{tikzpicture}
    \node[state] (c) at (0,0) {$C$};
    \node[state] (r) [right =of c] {$R$};


\end{tikzpicture}
$$
By intervening in the system, we are acting upon it from the outside. In doing so we disconnect those factors influencing the choice of clothing. The $do$ operator lets us express this external influence. However, if we don't have a $do$ operator there remains another option. Interventions can be represented by additional variables \cite{dawid2002}\footnote{This preprint has been corrected post-publication to include this citation of Dawid, as we were previously unaware of his work.}, so that we are no longer intervening in the system from outside. For example $do[C=true]$ might be represented by $A$ such that $p(C=true \mid A=true) = 1$ and $p(C \mid A=false) = p(C)$:

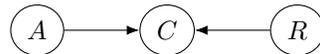
$$
\begin{tikzpicture}
    \node[state] (a) at (0,0) {$A$};

    \node[state] (c) [right =of a] {$C$};
    \node[state] (r) [right =of c] {$R$};

    \path (a) edge (c);
    \path (r) edge (c);

\end{tikzpicture}
$$
We can now represent that $p(R=true \mid C=true, A=true) = p(R=true) \neq p(R=true \mid C=true, A=false) = 1$. This expands the system to include an action by a specific actor, rather than accounting for interventions originating outside the system (as the $do$ operator does). 

\subsection{Emergent representation of interventions}
This does not entirely resolve our problem. Even if intervention is represented as a variable, that variable must still be explicitly defined before accurate induction can take place. It is an abstract notion which is presupposed. Variables are undefined in the context of definitions \ref{d1}, \ref{d2} and \ref{d3} for this very reason. Variables tend to be very abstract (for example, ``number of chickens'' may presuppose both a concept of chicken and a decimal numeral system), and the purpose (according to \cite{bennett2022a} and \cite{bennettmaruyama2022a}) of the formalism is to construct such abstractions via induction. It does so by formally defining reality (environment and cognition within that) using as few assumptions as possible \cite{bennett2023appendices}, in order to address symbol grounding \cite{harnad1990} and other problems associated with dualism. 
In this context, cause and effect are statements as defined in \ref{d2}. Returning to the example of Larry, instead of variables $A, C$ and $R$ we have a vocabulary $\mathfrak{v}$, and $c, r \in L_\mathfrak{v}$ which have a truth value in accordance with definition 2:
$$c \leftrightarrow \textit{``Larry put on a raincoat''} \text{ and } r \leftrightarrow \textit{``It rained''}$$
As before, assume we have concluded $p(r \mid c) = 1$ from passive observation, the naive interpretation of which is that we can make it rain by forcing Larry to wear a coat. However, the statement associated with this intervention is not \textit{just} $c = \textit{``Larry put on a raincoat''}$ but a third $a \in L$ such that: $$a \leftrightarrow \textit{``Larry put on a raincoat because we forced him to''}$$ 
$$
\begin{tikzpicture}
    \node[state] (a) at (0,0) {$a$};

    \node[state] (c) [right =of a] {$c$};
    \node[state] (r) [right =of c] {$r$};

    \path (a) edge (c);
    \path (r) edge (c);

\end{tikzpicture}
$$
Because we're now dealing with statements, and because statements are sets of declarative programs which are inferred rather than given, we no longer need to explicitly define interventions in advance. 
Statements in an implementable language represent sensorimotor activity, and are formed via induction \cite{bennett2022a,bennett2023appendices}.
The observation of $c$ is part of the sensorimotor activity $a$, meaning $c \subseteq a$ (if Larry is not wearing his raincoat, then it also cannot be true that we are forcing him to wear it). 
There is still no $do$ operator, however $i = a - c$ may be understood as representing the identity of the party undertaking the intervention. If $i \neq \emptyset$ then it is at least possible to distinguish intervention from passive observation, in the event that $a$ and $c$ are relevant (we still need explain under what circumstances this is true). 
Whether intervention and observation are indistinguishable depends upon the vocabulary $V$, the choice of which determines if $i=\emptyset$, or $i \neq \emptyset$ (the latter meaning that it is distinguishable).  
Thus interventions are represented, but only to the extent that the vocabulary permits.

\begin{definition}[intervention]\label{d9}
If $a$ is an intervention to force $c$, then $c \subseteq a$. Intervention is distinguishable from observation only where $c \subset a$.
\end{definition}
\subsection{When will induction distinguish intervention from observation?}
From \textbf{(prem. 1)} we have that choosing the weakest model maximises the probability of generalisation.
There are many combinations of parent and child task for which generalisation from child to parent is only possible by selecting a model that correctly distinguishes the effects of intervention from passive observation (a trivial example might be a task informally defined as ``predict the effect of this intervention''). It follows that to maximise the probability of generalisation in those circumstances the weakest model must distinguish between an intervention $a$ and what it forces, $c$, so long as \textbf{(prem. 2)} is satisfied as in def. \ref{d9}, s.t. $a \neq c$. 

\section{Awareness}
We have described how an intervention $a$ is represented as distinct from that which it forces, $c$. Induction will form models representing this distinction in tasks for which this aids completion. Now we go a step further.
Earlier we discussed $i = a - c$ as the identity of the party undertaking an intervention $a$. We might define a weaker identity as $k \subset i$, which is subset of any number of different interventions undertaken by a particular party. The $do$ operator assumes the party undertaking interventions is given, and so we might think of $k$ above as meaning ``me''. However, there is no reason to restrict emergent representations of intervention only to one's self. For example there may exist Harvey, who also intervenes to force $c$. It follows we may have $v$ such that $c \subset v$, and $v$ represents our observation of Harvey's intervention.  
$$
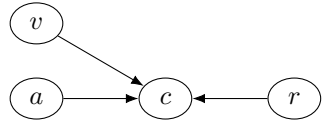
\begin{tikzpicture}
    \node[state] (a) at (0,0) {$a$};
    \node[state] (v) at (0,1) {$v$};
    \node[state] (c) [right =of a] {$c$};
    \node[state] (r) [right =of c] {$r$};

    \path (v) edge (c);
    \path (a) edge (c);
    \path (r) edge (c);

\end{tikzpicture}
$$
If $k \subseteq a - c$ can represent our identity as party undertaking interventions, it follows that $j  \subseteq  v - c$ may represent Harvey's. Both identities are to some extent context specific (another intervention may produce something other than $j$, or a subset of $j$, for Harvey), but these emergent identities still exist as a measurable quantity independent of the interventions with which they're associated.

\begin{definition}[identity]
If $a$ is an intervention to force $c$, then $k \subseteq a - c$ may function as an identity undertaking the intervention if $k \neq \emptyset$. 
\end{definition}
One's own identity is used to distinguish interventions from passive experiences to facilitate accurate inductive inference in an interactive setting. It follows from \textbf{(prem. 1)} that every object that has an impact upon one's ability to complete tasks must \textit{also} have an identity\footnote{Assuming interventions are distinguishable.}, because failing to account for the interventions of these objects would result in worse performance.

\subsection{Intent}
The formalism we are discussing originated as a mechanistic explanation of theory of mind called ``The Mirror Symbol Hypothesis'' \cite{bennettmaruyama2022a}, and of meaning in virtue of intent \cite{bennett2022a} (similar to Grice's foundational theory of meaning \cite{grice2007}). 
A statement is a set of declarative programs, and can be used as a goal constraint as is common in AI planning problems \cite{kautz92}. In the context of a task a model expresses such a goal constraint, albeit integrated with how that goal is to be satisfied \cite{bennett2022a,bennett2023appendices}. 
If one is presented with several statements representing decisions, and the situations in which they were made  (a task according to definition \ref{d3}), then the weakest statement with which which one can derive the decisions from the situations (a model) is arguably the \textit{intent} those decisions served \cite{bennett2022a}. 
Thus, if identity $k$ experiences interventions undertaken by identity $j$, then $k$ can infer something of the intent of $j$ by constructing a task definition and computing the weakest models \cite{bennett2022a}. 
This is a mechanistic explanation of how it is \textit{possible} that one party may infer another's intent. Assuming induction takes place according to definition \ref{d8}, then it is also \textit{necessary} to the extent that $k$ affect's $j$'s ability to complete tasks. Otherwise, $j$'s models would not account for $j$'s interventions and so performance would be negatively impacted.
However, a few interventions is not really much information to go on. Humans can construct elaborate rationales for behaviour given very little information, which suggests there is more to the puzzle. The Mirror Symbol Hypothesis argues that we fill in the gaps by projecting our own emergent symbols (either tasks or models, in this context) representing overall, long term goals and understanding onto others in order to construct a rationale for their immediate behaviour \cite{bennett2022a}, in order to empathise.

\subsection{How might we represent The Mirror Symbol Hypothesis?}
Assume there exists a task $\Omega$ which describes every decision $k$ might ever make which meets some threshold of ``good enough'' \cite{bennettmaruyama2022a,bennett2022a} at a given point in time. 
\begin{definition}[higher and lower level statements]
A statement $c \in L$ is higher level than $a \in L$ if $Z_{a} \subset Z_c$, which is written as $a \sqsubset c$.
\end{definition}
A model $m_\Omega \in M_\Omega$ is $k$'s ``highest level'' intent or goal (given the threshold), meaning $Z_\Omega = D_\Omega$.
Using $m_\Omega$ and $k$'s observation of decision $d$ made in situation $s$ by $j$  (the observation of which would also be a decision), $k$ could construct a lower level model $m_\omega \sqsubset m_\Omega$ such that $d \in Z_s \cap Z_{m_\omega}$. In other words, $m_\omega$ is a rationale constructed by $k$ to explain $j$'s intervention. 
Related work explores this in more depth \cite{bennett2022a,bennettmaruyama2022a}. For our purposes it suffices to point out that in combining emergent causality, identity, The Mirror Symbol Hypothesis \cite{bennettmaruyama2022a} and symbol emergence \cite{bennett2022a}, we have a mechanistic explanation of the ability to reason about one's own identity and intent, and that of others, in terms of interventions. Likewise the ability to predict how one's own intent is modelled by another is also of value in predicting that other's behaviour. In tasks of the sort encountered by living organisms, optimal performance would necessitate identity $k$ constructing a model of $j$'s model of $k$, and $j's$ model of $k's$ model of $j$ and so on to the greatest extent permitted by $\mathfrak{v}$ (the finite memory and any other limitations one's ability to represent predictions of predictions of predictions ad infinitum).

\subsection{Consciousness}
We have described a means by which an agent may be aware of itself, of others, of the intent of others and of the ability of others to model its own intent. By aware, we mean it has \textit{access} to and will function according to this information (access and functional consciousness, contextualising everything in terms of identities and their intent). 
Boltuc argues that phenomenal consciousness (characterised as first person functional consciousness) is explained by today's machine learning systems \cite{boltuc2012}. We would suggest his argument extends to our formalism, and in any case if qualia are a mechanistic phenomenon then they are already represented by the vocabulary of the implementable language. What is novel in our formalism is not just that it points out that causal inference may construct identity and awareness, but that it does so with a formulation that also addresses enactive cognition, symbol emergence and empathy \cite{bennettmaruyama2022a,bennett2022a}.

\subsubsection{Anthropomorphism:}
An implementation of what we have described would construct an identity for anything and everything affecting its ability to complete tasks - even inanimate objects like tools, or features of the environment. Intent would be ascribed to those identities, to account for the effect those objects have upon one's ability to satisfy goals. Though this might seem a flaw, to do anything else would negatively affect performance. Interestingly, this is consistent with the human tendency \cite{kotrschal2015} to anthropomorphise. We ascribe agency and intent to inanimate objects such as tools, the sea, mountains, the sun, large populations that share little in common, things that go bump in the night and so forth. 

\subsubsection{Fragmented identities:} It is also interesting to consider what this says of systems which are less than optimal (do not identify the weakest hypothesis), or which do not use a vocabulary which permits the construction of one identity shared by all of the interventions it undertakes. Such a thing might construct multiple unconnected identities for itself, and ascribe different intentions to each one. Likewise if the model constructs multiple identities for what is in fact the same object, it may hallucinate and hold contradictory beliefs about that object.


\printbibliography

@article{chalmers1995,
	publisher = {Imprint Academic},
	author = {David Chalmers},
	year = {1995},
	title = {Facing Up to the Problem of Consciousness},
	volume = {2},
	number = {3},
	pages = {200--19},
	journal = {Journal of Consciousness Studies}
}

@article{boltuc2012,
	author = {Boltuc, Piotr},
	title = {The Engineering Thesis in Machine Consciousness},
	journal   = {Techné: Research in Philosophy and Technology},
    volume    = {16},
    number = {2},
    pages = {187--207},
    year      = {2012}
}

@article{harnad1990,
    title = {The symbol grounding problem},
    journal = {Physica D: Nonlinear Phenomena},
    volume = {42},
    number = {1},
    pages = {335-346},
    year = {1990},
    %issn = {0167-2789},
    %doi = {https://doi.org/10.1016/0167-2789(90)90087-6},
    %url = {https://www.sciencedirect.com/science/article/pii/0167278990900876},
    author = {Stevan Harnad}
}

@article{ward2017,
author = {Ward, Dave and Silverman, David and Villalobos, Mario},
year = {2017},
month = {04},
pages = {},
title = {Introduction: The Varieties of Enactivism},
volume = {36},
journal = {Topoi}%,
%doi = {10.1007/s11245-017-9484-6}
}

@InProceedings{kautz92,
    author = {Henry Kautz and Bart Selman},
    title = {Planning as satisfiability},
    booktitle = {IN ECAI-92},
    year = {1992},
    pages = {359--363},
	address = {New York},
    publisher = {Wiley}
}

@book{bennett2023appendices,
%  author={Bennett, Michael Timothy},
%  publisher={GitHub},
%  title={Technical Appendices},
%  url={https://github.com/ViscousLemming/Technical-Appendices}
%  year={2023}
%}

@online{bennett2023a,
  author={Bennett, Michael Timothy},
  title={Computational Dualism and Objective Superintelligence}, 
  year={2023},
  url = {arxiv.org/abs/2302.00843}
}

@InProceedings{bennett2023b,
  author="Bennett, Michael Timothy",
editor="",
title="The Optimal Choice of Hypothesis Is the Weakest, Not the Shortest",
booktitle="Artificial General Intelligence",
year="2023",
publisher="Springer",
address="",
pages="42--51"
}

@InProceedings{bennett2023d,
author="Bennett, Michael Timothy",
editor="",
title="On the Computation of Meaning, Language Models and Incomprehensible Horrors",
booktitle="Artificial General Intelligence",
year="2023",
publisher="Springer",
address="",
pages="32--41"
}

@article{bennettmaruyama2022a,
  author={Bennett, Michael Timothy and Maruyama, Yoshihiro},
  journal={IEEE Transactions on Cognitive and Developmental Systems}, 
  title={Philosophical Specification of Empathetic Ethical Artificial Intelligence}, 
  year={2022},
  volume={14},
  number={2},
  pages={292-300}
}

@InCollection{heidegger2020,
	author       =	{Wheeler, Michael},
	title        =	{{Martin Heidegger}},
	booktitle    =	{The {Stanford} Encyclopedia of Philosophy},
	editor       =	{Edward N. Zalta},
	howpublished =	{\url{https://plato.stanford.edu/archives/fall2020/entries/heidegger/}},
	year         =	{2020},
	edition      =	{{F}all 2020},
	publisher    =	{Stanford University}
}

@InProceedings{bennett2022a,
    author="Bennett, Michael Timothy",
    %editor="Goertzel, B. and Ikl{\'e}, M. and Potapov, A.",
    title="Symbol Emergence and the Solutions to Any Task",
    booktitle="Artificial General Intelligence",
    year="2022",
    publisher="Springer",
    address="Cham",
    pages="30--40"%,
    %isbn="978-3-030-93758-4"
}

@article{block2002,
	title = {The Harder Problem of Consciousness},
	%doi = {10.2307/3655621},
	journal = {Journal of Philosophy},
	year = {2002},
	pages = {391},
	publisher = {Journal of Philosophy},
	author = {Ned Block},
	volume = {99},
	number = {8}
}

@article{franklin2008,
	title = {A Phenomenally Conscious Robot?},
	journal = {APA Newsletter on Philosophy and Computers},
	year = {2008},
	publisher = {APA},
	author = {S. Franklin, and B. J. Baars, and U. Ramamurthy.},
	number = {1}
}

@book{grice2007, 
    title={Studies in the Way of Words}, 
    publisher={Harvard University Press}, 
    author={Grice, Herbert P.}, 
	address = {Cambridge MA},
    year={2007}
}

@article{nickerson1998,
author = {Raymond S. Nickerson},
title ={Confirmation Bias: A Ubiquitous Phenomenon in Many Guises},
journal = {Review of General Psychology},
volume = {2},
number = {2},
pages = {175-220},
year = {1998}
%doi = {10.1037/1089-2680.2.2.175},
}

@article{bekinschtein2018,
  title={A retrieval-specific mechanism of adaptive forgetting in the mammalian brain},
  author={Bekinschtein, Pedro and Weisstaub, Noelia V and Gallo, Francisco and Renner, Maria and Anderson, Michael C},
  journal={Nature Communications},
  volume={9},
  number={1},
  pages={4660},
  year={2018},
  publisher={Nature Publishing Group UK London}
}

@article{berlin1990,
author = {Berlin, Sharon B.},
title = {Dichotomous and Complex Thinking},
journal = {Social Service Review},
volume = {64},
number = {1},
pages = {46-59},
year = {1990}
}

@article{kotrschal2015,
title = {The mind behind anthropomorphic thinking: attribution of mental states to other species},
journal = {Animal Behaviour},
volume = {109},
pages = {167-176},
year = {2015},
%issn = {0003-3472},
%doi = {https://doi.org/10.1016/j.anbehav.2015.08.011},
%url = {https://www.sciencedirect.com/science/article/pii/S0003347215003085},
author = {Esmeralda G. Urquiza-Haas and Kurt Kotrschal},
keywords = {animals, anthropomorphism, dual process theory, empathy, reasoning, social cognition}
}

@book{hutter2010,
    author = {Hutter, Marcus},
    title = {Universal Artificial Intelligence: Sequential Decisions Based on Algorithmic Probability},
    year = {2010},
    %isbn = {3642060528},
    publisher = {Springer-Verlag},
    address = {Berlin, Heidelberg}
}

@book{bennettmaruyama2021b,
  title={Intensional Artificial Intelligence: From Symbol Emergence to Explainable and Empathetic AI},
  author={Bennett, M. T. and Maruyama, Y.},
  %journal={arXiv preprint arXiv:2104.11573 [cs.AI]},
  publisher={Manuscript},
  %journal ={Manuscript},
  year={2021}
}

@book{ortega2021,
  %doi = {10.48550/ARXIV.2110.10819},
  %url = {https://arxiv.org/abs/2110.10819},
  author = {Ortega, Pedro A. and Kunesch, Markus and Delétang, Grégoire and Genewein, Tim and Grau-Moya, Jordi and Veness, Joel and Buchli, Jonas and Degrave, Jonas and Piot, Bilal and Perolat, Julien and Everitt, Tom and Tallec, Corentin and Parisotto, Emilio and Erez, Tom and Chen, Yutian and Reed, Scott and Hutter, Marcus and de Freitas, Nando and Legg, Shane},
  keywords = {Machine Learning (cs.LG), Artificial Intelligence (cs.AI), FOS: Computer and information sciences, FOS: Computer and information sciences},
  title = {Shaking the foundations: delusions in sequence models for interaction and control},
  journal = {Deepmind},
  year = {2021},
  address = {Deepmind}%,
  %edition = {1st}
  %copyright = {arXiv.org perpetual, non-exclusive license}
}

@book{pearl2018,
	author = {Pearl, Judea and Mackenzie, Dana},
	title = {The Book of Why: The New Science of Cause and Effect},
	year = {2018},
	publisher = {Basic Books, Inc.},
	address = {New York},
	edition = {1st}
}

@book{pearl2009, 
    place={Cambridge}, 
    edition={2}, 
    title={Causality}, %DOI={10.1017/CBO9780511803161}, 
    publisher={Cambridge Uni. Press}, 
    author={Pearl, Judea}, 
	address = {United Kingdom},
    year={2009}
}

@article{pearl1995,
    %ISSN = {00063444},
    %URL = {http://www.jstor.org/stable/2337329},
    author = {Judea Pearl},
    journal = {Biometrika},
    number = {4},
    pages = {669--688},
    publisher = {[Oxford University Press, Biometrika Trust]},
    title = {Causal Diagrams for Empirical Research},
    urldate = {2022-07-06},
    volume = {82},
    year = {1995}
}

@article{dawid2002,
 ISSN = {03067734, 17515823},
 URL = {http://www.jstor.org/stable/1403901},
 author = {A. P. Dawid},
 journal = {International Statistical Review / Revue Internationale de Statistique},
 number = {2},
 pages = {161--189},
 publisher = {[Wiley, International Statistical Institute (ISI)]},
 title = {Influence Diagrams for Causal Modelling and Inference},
 urldate = {2024-02-22},
 volume = {70},
 year = {2002}
}

\end{document}